# Simulation of Genetic Algorithm: Traffic Light Efficiency

# Senior Research Paper

# By: Eric Lienert

**Abstract:**

Traffic is a problem in many urban areas worldwide.  Traffic flow is dictated by certain devices such as traffic lights.  The traffic lights signal when each lane is able to pass through the intersection. Often, static schedules interfere with ideal traffic flow.  The purpose of this project was to find a way to make intersections controlled with traffic lights more efficient.  This goal was accomplished through the creation of a genetic algorithm, which enhances an input algorithm through genetic principles to produce the "fittest" algorithm.  The program was comprised of two major elements: coding in Java and coding in Simulation of Urban Mobility (SUMO), which is an environment that simulates real traffic.  The Java code called upon the SUMO simulation via a command prompt which ran the simulation, received the output, altered the algorithm, and looped.  The SUMO component initialized a simulation in which a 1x1 street layout was created, each intersection with its own traffic light.  Each loop enhanced the input algorithm by altering the scheduling string (dictates the light changes).  After the looped simulations were executed, the data was then analyzed.  This was accomplished by creating an algorithm based upon "regular" practice – timed traffic lights – and comparing the output which was comprised of the total time it took for all vehicles to exit the system and the average time it took each individual vehicle to exit the system.  These different variables: the time it took the average vehicle to exit the system and total time for all vehicles to exit the system, where then graphed together to provide a visual aid.  The genetic algorithm did improve traffic light and traffic flow efficiency in comparison to traditional scheduling methods.

**Acknowledgements:**

                              Mrs. Yarbrough
                              Nath Tumlin
                              Dr. Gray
                              Classmates
                              Family
                              Brother

**Biography:**

Name: Eric Lienert

**Table of Contents**

## Chapter 1: The Problem

**Introduction and Problem:**

Everyone encounters problems with traffic throughout their busy lives. This problem is due to the fact that the off-ramps from interstates lead to intersections controlled by traffic lights. These traffic lights are usually on timers or weight plates and thus, do not optimize traffic flow. Therefore, to reduce traffic, traffic lights must become more efficient. The purpose of this study is to create a genetic algorithm which performs that very purpose. A genetic algorithm is an algorithm which uses the tenants of heredity and incorporates them into a self-learning program that will create the "best" algorithm from a given set of parameters. Everyone's commute to work will improve as a direct consequence through this effort. This study will use collected traffic data to optimize the traffic signals and provide real-world results. Traffic light scheduling is inflexible and produces traffic problems as a result. Therefore, a flexible algorithm must be utilized to adapt to a changing environment; an algorithm which is efficient in most any situation. A traffic light scheduling algorithm that takes a certain set of given parameters (number of cars, speed, acceleration, etc.) and creates an efficient schedule would yield the best results. As such, a genetic algorithm was created in an attempt to solve all of these complexities and issues in Java code. A simulation portion was needed to yield data for the genetic algorithm

**Hypothesis, Variables, and Limitations**

      The created program has the potential to lead to an increased efficiency of traffic signals and thus reduce the amount of traffic. Due to the nature of the program created, the data will

remain constant but the output will be ever changing, keeping in line with the idea of a genetic algorithm; ever changing. Thus, all the procedures will be carried out in a constant environment due to the nature of the simulation. This process will be performed as many times as the Java environment allows the genetic algorithm to create the "fittest" schedule.

**Assumptions:**

    Simulation of Urban Mobility (SUMO) is a language used to create traffic simulations. It is assumed that the SUMO environment will accurately depict the real life situation of traffic flow. The data that is collected from the real-world is assumed to be accurate.

**Statistical Analysis:**

    The statistical procedure required for this project is the relationship between the time of the genetic algorithm versus the time of the standard traffic light. This relationship was graphed to present a visual representation of the association as well. This relationship between the two displayed the relative efficiency of the newly created scheduling algorithm versus the standard non-genetic algorithm. Thus, this was a very accurate test as to whether the program created a more efficient algorithm in comparison.

**Definitions:**

- *Genetic Algorithm* - an algorithm which applies genetic principles to produce the "most fit" algorithm

- *SUMO* – a traffic simulation program, the Simulation of Urban Mobility

**Chapter 2: Literature Review**

Traffic is a problem most everyone has faced, especially in developed areas due to it being the main vein of transportation for most people.  Traffic has caused accidents, tardiness, and general aggression.  Many factors affect traffic patterns and traffic jams.  The most potent and problematic factors have been identified as the general public's mindset, accidents, and the traffic flow system itself.  The mindset of the public is that closely following the car in front and aggressive driving helps to mitigate the problem.  This, in fact, is basically the opposite of how traffic will revert back to normal traffic flow.  In that, if one were to leave a large amount of room in front of them and move at a constant speed, the traffic would change to a more standard pace.  Accidents are a problem that caused many traffic slow-downs.  Although they cause some of the worst traffic jams, accidents cannot truly be fought against on a large scale as they are caused by individuals.  Lastly, the set-up of the traffic grid further compound the problem.  These ailments cannot be targeted on the highway as there is no constrictive element that slows cars other than the cars themselves.  Therefore, this study aims to amend the traffic system, specifically the traffic lights.

Traffic lights conduct the flow of traffic on many roads.  Since they were originally conceptualized so long ago, there are out of date and in need of a new system.  A highly promising algorithm resulted from the idea of genetic algorithms.  Genetic algorithms are mathematical instructions that are fed a certain parameter and what is qualified as "good".  The algorithm then, incrementally, finds the "best" algorithm.  If the traffic lights were fixed, it would allow traffic to flow off of the interstates and onto smaller roads; effectively dividing the traffic flow so that it would run more smoothly for all individuals.  Thus, this study aimed towards creating a genetic algorithm that would dictate the scheduling of traffic lights to help traffic move more efficiently.

Therefore, to more specifically pin point the problem, information was collected in respect to how the traffic lights operated, different hypotheses of traffic patterns, and these all tie together. Since some causes of traffic are now known, it can be determined that it the interaction of the scheduling, and thus the devices, and of the individuals that causes the problems of traffic to occur. Many individuals implement the practice of tailgating in which they follow the car in front of them very closely, leaving little room between the two. As more and more cars commit this, the effect is amplified until it reaches the back of the traffic. One comparison to how traffic works is the slinky image in which the whole of traffic is a giant slinky. The slinky constricts and relaxes at different points, the constricted parts symbolizing high traffic concentrations whereas the relaxed portions represent the freely flowing traffic.

The largest portion of the project lies within the code that functions behind the scenes. It is wholly based upon the genetic algorithm. This specific algorithm is based upon meiosis and the rules that dictate the transition of genes. These rules allow the genetic algorithm to become "better" after each "generation" is created. Thus, after executing the script many times, the algorithm's fitness increases by a large margin. Therefore, it is very important to fully understand the background information on this subject since it is so fundamental to this project.

This project also has a large portion vested in traffic lights. Traffic lights are scheduling devices used to dictate traffic flow through intersections. Therefore, it is necessary for this study to understand the inner workings of how traffic lights function.   Additionally, it is also important to better comprehend scheduling in computing in general.   Traffic lights are usually set by a timer or by some sort of mechanism (weight or magnetic) underground that indicates when cars are near.  These resources allow the study to obtain more specific information in relation to traffic lights, the operation of traffic lights, and how they operate mechanically.

Lastly, it is important for this study to better understand how traffic occurs and why it develops.  This allows the study to create a better algorithm program because the program is dependent upon traffic and how it flows.  Thus, resources dealing with traffic are crucial to the project as well.

**Eliminating unexplained traffic jams**

Horn investigates the problem with rush hour traffic in which the regular rush hour traffic ends up becoming a traffic jam.  He created an algorithm which should be implemented into future cars much like cruise control.  This algorithm would use sensors to establish the distance between the car in front and behind and maintain a distance half way between them, as such, Horn described his algorithm as "bilateral".  Horn hypothesizes that the reason for these unexplained traffic jams is due to the fact that cars further from the traffic light take a longer time to start after the light turns green as opposed to cars close to the light which start right away.  One problem described in Horn's paper is that many cars need to implement his "bilateral" system which involves radar detection devices which are relatively expensive pieces of equipment.  Lastly, another problem involves the self-absorbed nature of drivers in general, in that, each driver worries about his/herself as opposed to the traffic community as a whole.

**Dynamic Traffic Light Sequence Algorithm Using RFID**

This project attempted to create a more efficient schedule for traffic lights by feeding real-time data via RFID technology which using light beams and sensors to collect the number of cars passing through the intersection. These sensors allowed the intersections to actively adapt to a changing environment which led to a well-rounded scheduling of cars. The problem with this approach was similar to the former investigation, as the sensors had to be installed in order to be effective which requires a large sum of capital.

A self-adapting genetic algorithm for project scheduling under resource constraints

Hartmann used a genetic algorithm to create a self-learning program that would best adapt itself to given parameters. The experiment was successful as the algorithm did indeed become more efficient as it progressed through its iterations.

**Chapter 3: Materials and Methods**

Introduction

A java program was created implementing a genetic algorithm and using Simulation of Urban Mobility to create a simulation environment.  This program has created a more efficient traffic light scheduling algorithm for the given circumstances created in the SUMO environment.

*Materials*

1. A computer
2. Access to the internet
3. Administrator privilege
4. Dr. Java IDE
5. Java Compiler
6. SUMO
7. Python 2.

*Methods*

Java was installed.  Dr Java was installed.  SUMO was installed.  Python was installed.  A genetic algorithm was created in Java code after an algorithm class, object class (light in this project), and a population of the objects class called a Populace.  An executable command prompt within Java was created to call upon SUMO code in which SUMO fed information back into Java.  A SUMO simulation was created to provide raw data after the node, routes, and edge files were created. In addition, traffic lights were created at each of the intersections. A loop was created in Java to execute this series of steps many times to make the program "evolve".

*Data Collection*

This loop was run multiple times, with each iteration passing certain "genes" onto the next generation. The "best" light scheduling string was returned in Java. Java then printed the scheduling string in English vernacular.

*Appendix (Blank Charts)*

| Car Number | Entrance Time | Departure Time | Total Time |
|---|---|---|---|

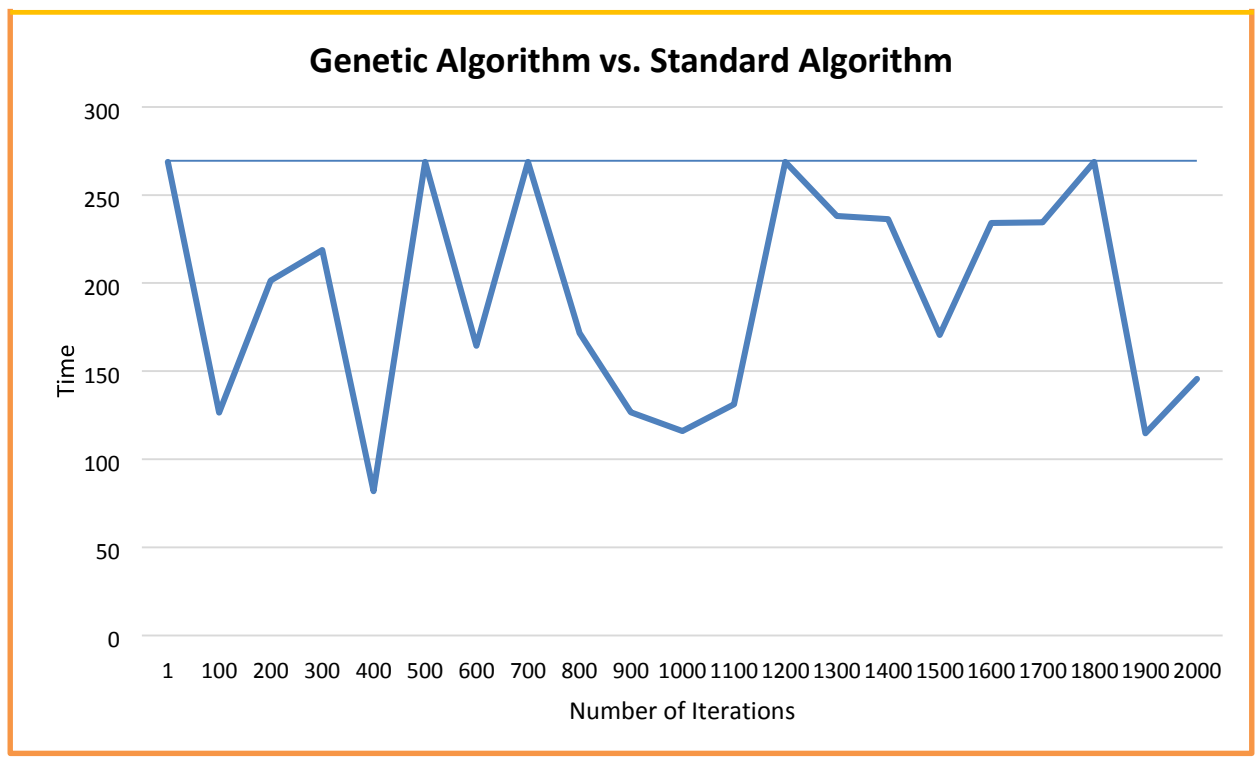

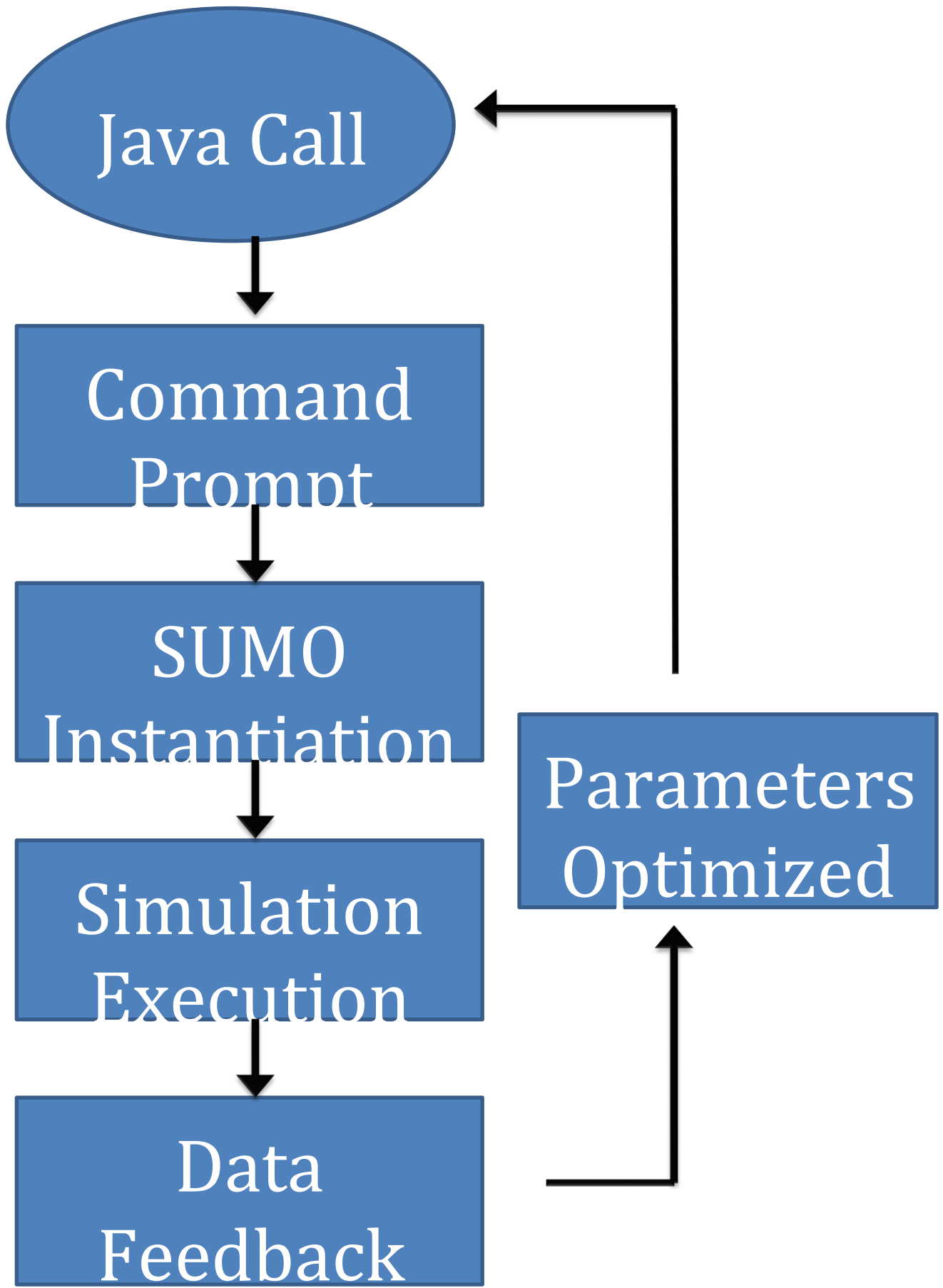

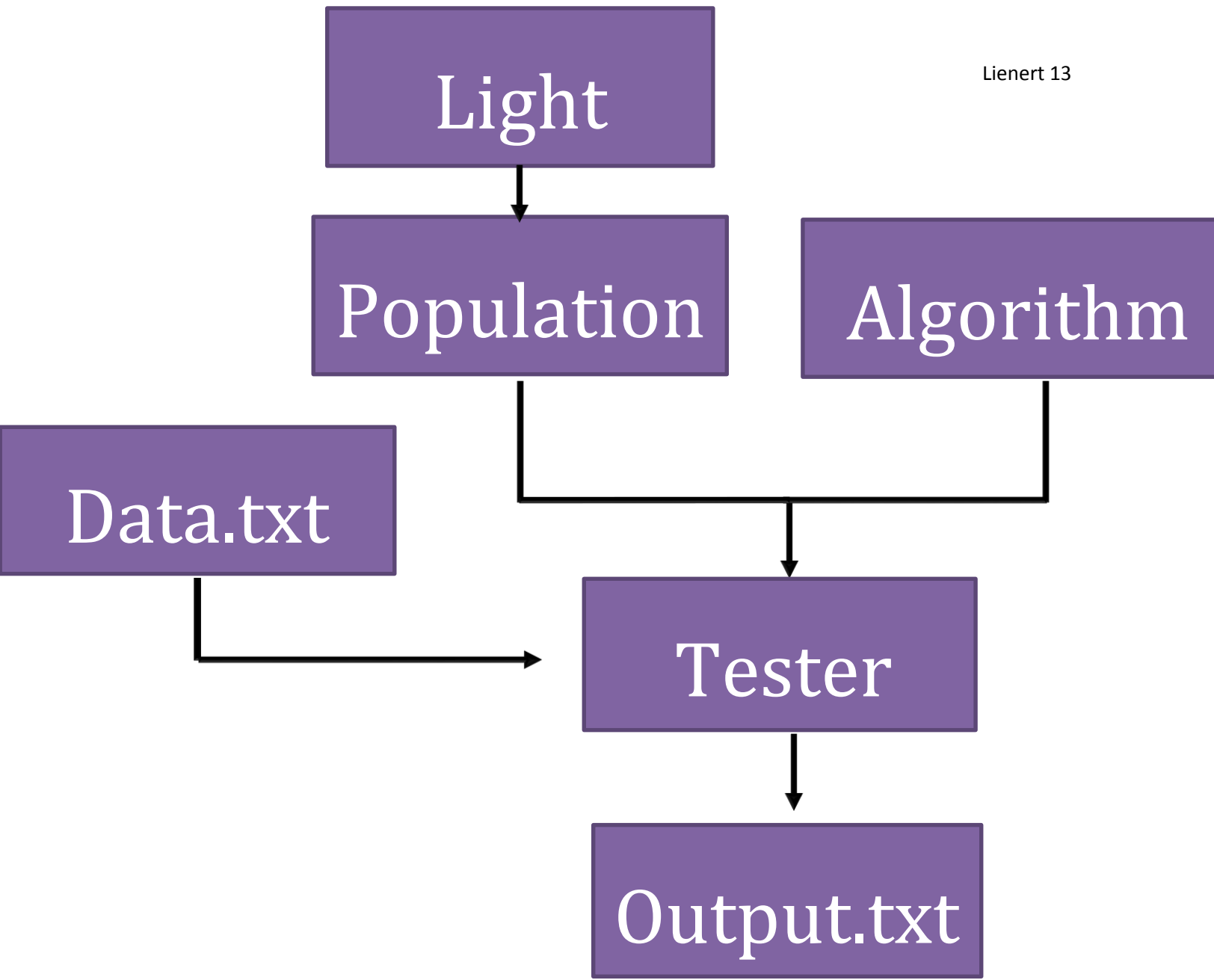

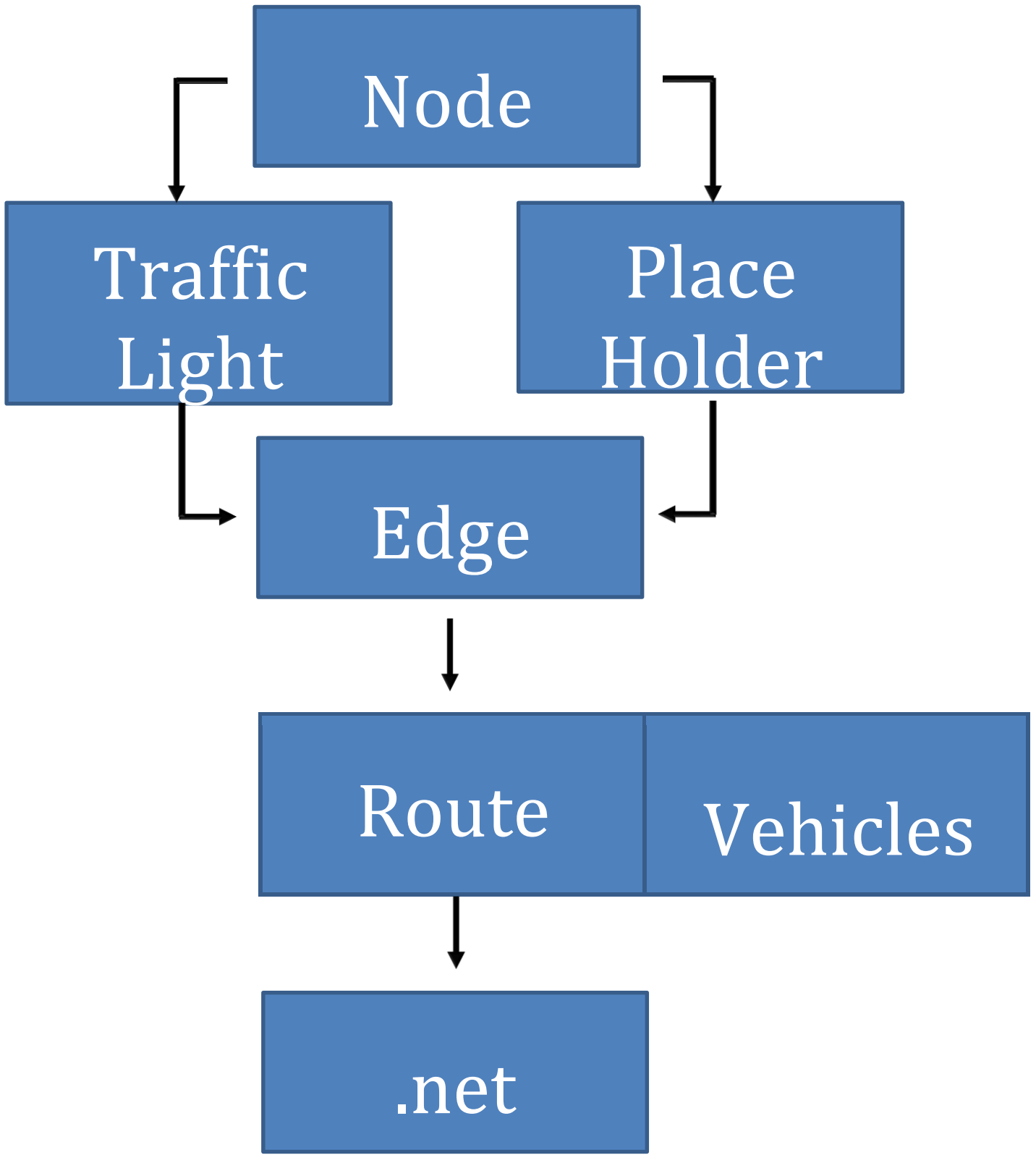

**Chapter 4: Results**

This table described the amount of time it took all vehicles to exit the system given the parameters set by Algorithm A and Algorithm B respectively. The algorithm which exited all vehicles in the least amount of time was deemed "better" and would thus pass traits onto the next iteration of algorithms. Because this involved creating a more efficient process, all that was needed was a graph to describe the change in efficiency. The genetic algorithm was found to increase efficiency by 92.1% after 2000 iterations. This shows the potential for the genetic algorithm to improve the efficiency even though the function was oscillating, suggesting that the genetic algorithm parameters were too constrictive.

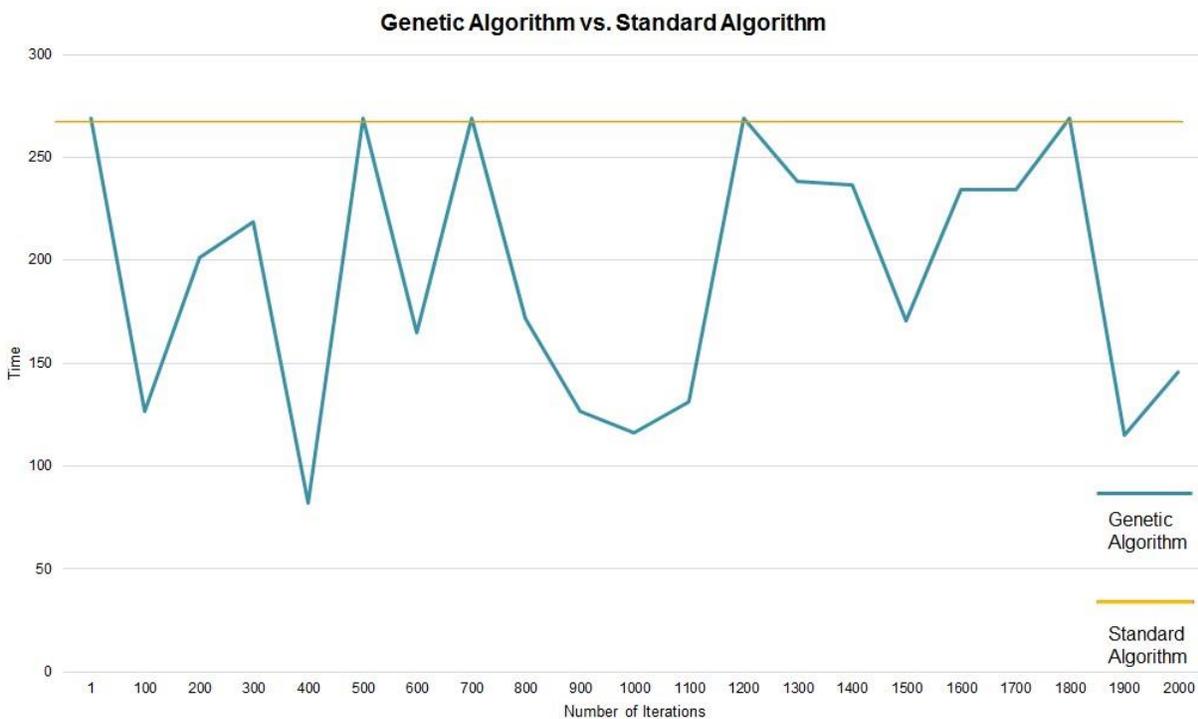

## Chapter 5: Conclusions

This project showed the practical application of a genetic algorithm to increase the efficiency of traffic lights to improve traffic conditions.  The genetic algorithm was proven to improve the efficiency and thus shown to work.  Traffic is a problem in many urban areas worldwide.  Traffic flow is dictated by certain devices such as traffic lights.  These traffic lights signal when each lane is able to pass through the intersection.  Often times, static schedules interfere with ideal traffic flow.  The purpose of this project was to find a way to make intersections controlled with traffic lights more efficient.  This goal was accomplished through the creation of a genetic algorithm.  A genetic algorithm enhances an input algorithm through genetic principles to produce the "fittest" algorithm.  The program was comprised of two major elements: coding in Java and coding in Simulation of Urban Mobility (SUMO).  SUMO is an environment which simulates real traffic.  The Java code called upon the SUMO simulation via a command prompt which ran the simulation, received the output, altered the algorithm, and looped many times over.  The SUMO component initialized a simulation in which a 1x1 street layout was created, each intersection with its own traffic light.  Each loop enhanced the input algorithm by altering the scheduling string (dictates the light changes).  After the looped simulations were executed, the data was then analyzed.  This was accomplished by creating an algorithm based upon "regular" practice – timed traffic lights – and comparing the output which was comprised of the total time it took for all vehicles to exit the system and the average time it took each individual vehicle to exit the system.  These different variables: time it took the average vehicle to exit the system and total time for all vehicles to exit the system, where then graphed together to provide a visual aid.  The genetic algorithm did improve traffic light and traffic flow efficiency in comparison to traditional scheduling methods. Thus, this project can be applied to a larger traffic system to help reduce traffic. This allowed for the data to apply to real-world applications. Genetic Algorithm was not given full array of parameters to change, limiting the results. SUMO was found to be sensitive in that a single change in the traffic light string could cause a dramatic shift in traffic time.

*Future Research*

- Expand setting to include multiple traffic lights to allow the development of an algorithm which makes the lights function together

- Implement a real-time data collecting device to allow the algorithm to function real time and adapt to changing conditions

- Collect real life data to eliminate confounding factors